%% file: example_paper.tex
\documentclass{article}

\usepackage{microtype}
\usepackage{graphicx}
\usepackage{booktabs} %
\input{math_commands.tex}

\usepackage{subcaption}
\usepackage{diagbox}
\usepackage{array}
\usepackage{multirow}
\usepackage{makecell}

\newcommand{\DMSp}{\(\text{DMS}_{\text{p}}\)}
\newcommand{\DMSnp}{\(\text{DMS}_{\text{np}}\)}
\newcommand{\DMSpn}{\(\text{DMS}_{\text{p-}}\)}

\usepackage{hyperref}

\usepackage[accepted]{icml2024}

\usepackage{amsmath}
\usepackage{amssymb}
\usepackage{mathtools}
\usepackage{amsthm}

\usepackage[capitalize,noabbrev]{cleveref}

\theoremstyle{plain}

\theoremstyle{definition}

\theoremstyle{remark}

\usepackage[textsize=tiny]{todonotes}

\icmltitlerunning{Submission and Formatting Instructions for ICML 2024}

\begin{document}

\twocolumn[
    \icmltitle{Differentiable Model Scaling using Differentiable Topk}

    \begin{icmlauthorlist}
        \icmlauthor{Kai Liu}{lab}
        \icmlauthor{Ruohui Wang}{lab}
        \icmlauthor{Jianfei Gao}{lab}
        \icmlauthor{Kai Chen}{lab}
    \end{icmlauthorlist}

    \icmlaffiliation{lab}{Shanghai AI Laboratory, Shanghai, China}
    \icmlcorrespondingauthor{Kai Liu}{liukai@pjlab.org.cn}
    \icmlcorrespondingauthor{Kai Chen}{chenkai@pjlab.org.cn}

    \icmlkeywords{Neural Architecture Search, Model Scaling, Machine Learning, ICML}

    \vskip 0.3in
]

\printAffiliationsAndNotice{}  %

\begin{abstract}
    Over the past few years, as large language models have ushered in an era of intelligence emergence, there has been an intensified focus on scaling networks. Currently, many network architectures are designed manually, often resulting in sub-optimal configurations. Although Neural Architecture Search (NAS) methods have been proposed to automate this process, they suffer from low search efficiency.
    This study introduces \emph{Differentiable Model Scaling (DMS)}, increasing the efficiency for searching optimal width and depth in networks.
    DMS can model both width and depth in a direct and fully differentiable way, making it easy to optimize.
    We have evaluated our DMS across diverse tasks, ranging from vision tasks to NLP tasks and various network architectures, including CNNs and Transformers.
    Results consistently indicate that our DMS can find improved structures and outperforms state-of-the-art NAS methods.
    Specifically, for image classification on ImageNet, our DMS improves the top-1 accuracy of EfficientNet-B0 and Deit-Tiny by 1.4\% and 0.6\%, respectively, and outperforms the state-of-the-art zero-shot NAS method, ZiCo, by 1.3\% while requiring only 0.4 GPU days for searching.
    For object detection on COCO, DMS improves the mAP of Yolo-v8-n by 2.0\%.
    For language modeling, our pruned Llama-7B outperforms the prior method with lower perplexity and higher zero-shot classification accuracy.
    We will release our code in the future.
\end{abstract}

\section{Introduction}

In recent years, large models such as GPTs \cite{radford2018gpt} and ViTs \cite{dosovitskiy2020vit} have showcased outstanding performance. Notably, the emergent intelligence of GPT4 \cite{openai2023gpt4} has underscored the importance of scaling networks as a critical pathway toward achieving artificial general intelligence (AGI).
To support this scaling process, we introduce a general and potent method to determine the optimal width and depth of a network during its scaling.

Currently, the structure design of most networks still relies on human expertise. It typically demands significant resources to tune structural hyperparameters, making it challenging to pinpoint the optimal structure. Meanwhile, \emph{Neural Architecture Search (NAS)} methods have been introduced to automate network structure design. We classify NAS methods into two categories based on their search strategies: \emph{stochastic search methods} \cite{xie2022scalenet,liu2022modelamplification,tan2019efficientnet} and \emph{gradient-based methods} \cite{liu2018darts,wan2020fbnetv2,guo2021jointpruning}.

The stochastic search methods involve sampling numerous sub-networks to compare performance. However, these methods are limited to low search efficiency due to the sample-evaluate cycle, leading to reduced performance and increased search costs.

\begin{figure*}[t]
    \begin{center}
        \includegraphics[width=0.75\textwidth]{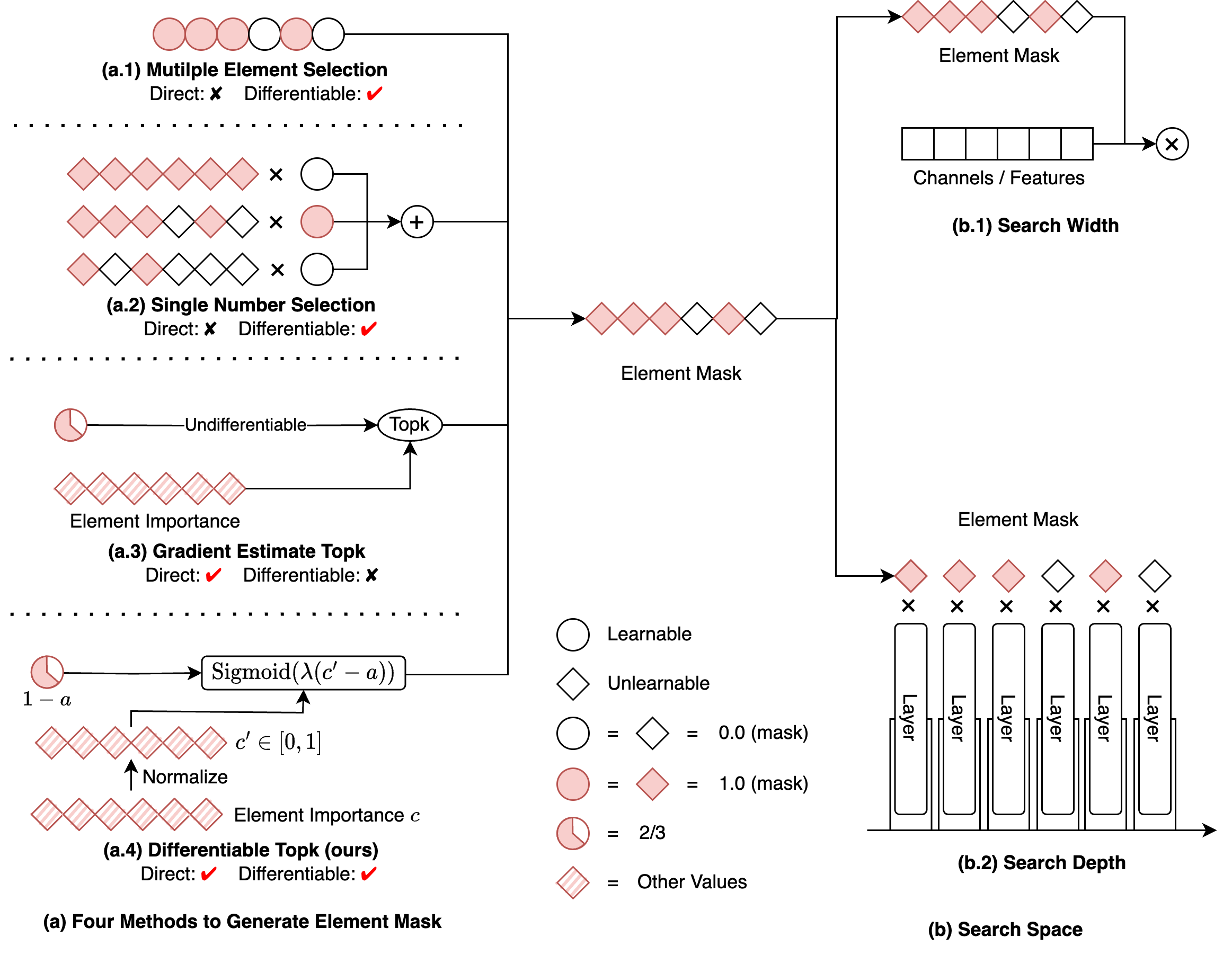}
        \caption{
            Different Gradient-based Modeling Strategies for Width and Depth. For all strategies, they use learnable parameters to generate an element mask to select width elements or depth elements.  SubFigure \textbf{(a)} illustrates four methods to generate the element mask, while \textbf{(b)} shows how the mask is used to search width and depth.
            \textbf{(a.1)} Multiple Element Selection: The element count is transformed into a multiple-element selection.
            \textbf{(a.2)} Single Number Selection: The element count is transformed into a selection from multiple numbers.
            \textbf{(a.3)} Gradient Estimate Topk: The element count is directly modeled yet non-differentiable.
            \textbf{(a.4)} Our Differentiable Topk: The element count is directly modeled and is fully differentiable.
            ``Direct'' means that the learnable parameters directly model the structural hyperparameters, while ``Differentiable'' means that the gradient of the learnable parameters can be computed in a fully differentiable manner.
        }
        \label{fig:compare}
    \end{center}
\end{figure*}

Unlike stochastic search methods, gradient-based methods employ gradient descent to optimize structural parameters, enhancing their efficiency and making them more adept at balancing search costs with ultimate performance.
However, a significant challenge persists:
\emph{how to model structural hyperparameters in \textbf{a direct and differentiable manner}}. Prior methods have struggled to meet this challenge, resulting in diminished performance and increased costs.
Specifically, we group prior methods into three categories based on their modeling strategies: (1) \emph{multiple element selection}, (2) \emph{single number selection}, and (3) \emph{gradient estimate topk}.
Specifically, when searching for the number of channels in a convolutional layer, multiple element selection methods \cite{lipas,guo2021gdp} model the channel number as multiple selections of channels, as shown in Figure \ref{fig:compare} (a.1). They introduce a much larger search space of element combinations.
Single number selection methods \cite{wan2020fbnetv2} model the channel number as a single selection from multiple numbers, as shown in Figure \ref{fig:compare} (a.2). It ignores the order relationship among these numbers.
Gradient estimate topk approaches \cite{guo2021jointpruning,gao2022ddnp,ning2020dsa} attempt to model width and depth directly, as shown in Figure \ref{fig:compare} (a.3). However, they are not differentiable, necessitating the development of different gradient estimation methods. As a result, these methods lack stability and are difficult to optimize.

Regrettably, all the above strategies fall short of modeling structural hyperparameters in a clear-cut and fully differentiable fashion.
To address the aforementioned challenge, we introduce  \emph{a fully differentiable topk operator}, which can seamlessly model depths and widths in a direct and differentiable manner. Notably, each differentiable topk operator has a single learnable parameter, representing either a depth or width structural hyperparameter. It can be optimized based on guidance from both task loss and resource constraint loss.
Our method stands out in terms of high optimization efficiency when contrasted with existing gradient-based approaches.

Based on our differentiable topk, we develop a \emph{Differentiable Model Scaling (DMS)} algorithm to search for networks' optimal width and depth.
To validate the efficacy and efficiency of our approach, we rigorously tested it across various tasks, including vision tasks and NLP tasks, and different architectures, including CNNs and Transformers.
Thanks to the high search efficiency of our differentiable topk, DMS achieves better performance or much lower search costs than prior SOTA methods.

Overall, our contributions are as follows:
\begin{itemize}
    \item We introduce a differentiable topk operator, which is easy to optimize as it can model structural hyperparameters in a direct and differentiable manner.
    \item We develop a Differentiable Model Scaling (DMS) algorithm based on our differentiable topk to search for networks' optimal width and depth.
    \item We evaluate our DMS across various tasks and architectures. For example, DMS outperforms the state-of-the-art zero-shot NAS method, ZiCo, by 1.3\% while requiring only 0.4 GPU days for searching. DMS costs fewer than a fraction of dozens of the search costs of the one-shot NAS method, ScaleNet, and the multi-shot NAS method, ModelAmplification with comparable performance. Besides, our method is a widely applicable method, which improves the mAP of Yolo-v8-n by 2.0\% on COCO and improves the zero-shot classification accuracy of pruned Llama-7B.
\end{itemize}

\section{Related Work}

The \emph{width} and \emph{depth} of networks are critical aspects of model architecture design.
A multitude of methodologies have been proposed to automate this process, notably Neural Architecture Search (NAS) \cite{zoph2016rlnas,liu2018darts} and \emph{model structure pruning} \cite{li2020eagleeye,lipas}. NAS algorithms typically aim to design models automatically from scratch, while model structure pruning approaches focus on compressing pretrained models to enhance their efficiency. Despite their contrasting methodologies, both approaches contribute to the search for model structure.

These search methods can generally be categorized into two groups based on their search strategies: stochastic search methods \cite{zoph2016rlnas,xie2022scalenet,liu2022modelamplification} and gradient-based methods \cite{liu2018darts,guo2021jointpruning}.
In the following sections, we will introduce these methods and compare them with ours.

\subsection{Stochastic Search Methods}

Stochastic search methods usually operate through a cyclical process of sampling and evaluation. At each step, they sample models with different structures and then evaluate them. This strategy is versatile as it can handle both contiguous and discrete search spaces.
However, a significant downside is its low search efficiency, leading to high resource consumption and suboptimal performance.
Specifically, stochastic search-based methods can be divided into three groups: \emph{multi-shot NAS}, \emph{one-shot NAS}, and \emph{zero-shot NAS}.
Multi-shot NAS \cite{tan2019efficientnet,liu2022modelamplification} requires the training of multiple models, which is time-consuming. For instance, EfficientNet \cite{tan2019efficientnet} uses over 1714 TPU days for searching.
One-shot NAS \cite{xie2022scalenet,cai2019once} requires training a large supernet, which is also resource-intensive. For example, ScaleNet \cite{xie2022scalenet} uses 379 GPU days for training a supernet.
Zero-shot NAS \cite{li2023zico,lin2021zen} reduces the cost by eliminating the need to train any model. However, its performance has not yet met the desired standard.

\subsection{Gradient-based Methods}

Gradient-based structure search methods \cite{liu2018darts,guo2021jointpruning} employ gradient descent to explore the structure of models. Generally, these methods are more efficient than their stochastic search counterparts.
The critical aspect of gradient-based methods is how to use learnable parameters to model structural hyperparameters and compute their gradients.
Ideally, the learnable parameters should directly model structural hyperparameters, and their gradients should be computed in a fully differentiable manner.
However, prior methods have struggled to meet these two criteria in modeling the width and depth of networks.
We group them into three categories: (1) multiple element selection, (2) single number selection, and (3) gradient estimate topk.
The first two categories model structural hyperparameters indirectly, while the third category is not differentiable and requires gradient estimation.

Multiple element selection methods \cite{lipas} model the number of elements as multiple selections from elements (e.g., channel selection), as shown in Figure \ref{fig:compare} (a.1).
Similarly, Single number selection methods \cite{wan2020fbnetv2} model element quantity as a single choice from multiple numbers, as shown in Figure \ref{fig:compare} (a.2).
Both them model structural hyperparameters in indirect and inaccurate ways and introduce much more learnable structural parameters, making optimization hard. Naturally, They result in low performance.

Gradient estimate topk approaches \cite{guo2021jointpruning,gao2022ddnp,ning2020dsa} attempt to model width and depth directly, as shown in Figure \ref{fig:compare} (a.3). However, they are not differentiable, necessitating the development of different gradient estimation methods. As a result, these methods lack stability and are also difficult to optimize.

To improve the optimization efficiency for structure search, we introduce a new differentiable topk that can model width and depth directly and is fully differentiable.
We compare our method and these search methods that are mentioned above in \hyperref[sec:ex_search]{Section 4.1}, The results show that our method is much more efficient and effective.

\section{Method}

In this section, we will detail our Differentiable Model Scaling (DMS) in two steps.
First, we introduce our differentiable topk, which models structural hyperparameters directly in a fully differentiable manner.
Second, we explain how to use our differentiable topk to construct our DMS algorithm.

\subsection{Differentiable Top-k}

Suppose there is a structural hyperparameter denoted by \( k \), representing the number of elements, such as \( k \) channels in a convolutional layer or \( k \) residual blocks in a network stage. \(k\) has a maximal value of \( N \).
We use \(\vc \in \mathbb{R}^N \) to represent the \emph{element importance}, where a larger value indicates a higher importance.
The objective of our differentiable topk is to output a soft mask \(\vm \in [0,1]^N \) to indicate the selected elements with top \(k\) importance scores.

Our topk operator uses a learnable parameter \(a\) as a threshold to select elements whose importance values are larger than \(a\).
\(a\) is able to model the number of elements \(k\) directly, as \(k\) can be seen as a function of \(a\), where \( k=\sum^N_{i=1}{1[c_i>a]}\).
\(1[A]\) is an indicator function, which equals 1 if the A is true and 0 otherwise. We use \(c_i\) to represent the importance of the \(i\)-th element.
We denote our topk as a function \(f\) as follows:

\begin{align}
    m_i = f(a) \approx \begin{cases}
                           1 & \text{if } c_i > a \\
                           0 & \text{otherwise}
                       \end{cases}
\end{align}

In prior methods, \(f\) is usually a piecewise function, which is not smooth and not differentiable, and the gradient of \(a\)  is computed by estimation.
We argue the biggest challenge to employing a fully differentiable \(f\) with respect to \(a\) is that the channel importance is distributed unevenly.
Specifically, uneven distribution causes the importance difference between two neighboring elements, ordered by importance value, to vary significantly. Supposed \(a\) is updated by a fixed value in each iteration, when the difference is large, a lot of steps are needed for \(a\) to go across these two elements.
When the difference is small, \(a\) can cross many elements in one step.
Therefore, optimizing \(a\) in a fully differentiable manner is too hard when element importance is uneven.

To address this challenge, we employ an \emph{importance normalization} process to forcefully convert the unevenly distributed importance to evenly distributed values, making the topk function smooth and easy to optimize in a differentiable way.
To sum up, our differentiable topk has two steps: importance normalization and \emph{soft mask generation}.

\subsubsection{Importance Normalization}

We normalize all element importance by mapping them to evenly distributed values from 0 to 1, based on the following:

\begin{align}
     & c_i' = \frac{1}{N}\sum^N_{j=1}{1[c_i>c_j]}.
    \label{eq:normalize}
\end{align}

The normalized element importance is denoted by \(\vc'\). \(1[A]\) is the same indicator function as above. Any two elements in \(\vc\) are supposed to be different, which is usually the case in practice.
Notably, although \(\vc'\) is evenly distributed from 0 to 1, \(\vc\) can follow any distribution.

Intuitively, \(c'_i \) indicates the portion of \(\vc \) values smaller than \(c_i \).
Besides, the learnable threshold \(a\) also becomes meaningful, representing the pruning ratio of elements. \(k\) can be computed by \(k=\lfloor(1-a)N\rceil\), where \( \lfloor \, \rceil \) is a round function.
\(a\) is limited to the range of \([0,1]\), where \(a=0\) indicates no pruning and \(a=1\) indicates pruning all elements.
Our ablation study in \hyperref[sec:abla_normal]{Appendix A.3.1} demonstrates that the importance normalization is critical to the performance of our topk.

\subsubsection{Soft Mask Generation}
\label{sec:mask}

After the normalization, it's easy to generate the soft mask \( \vm \) using a smooth and differentiable function based on the relative size of pruning ratio \(a\) and normalized element importance \( \vc' \).

\begin{align}
     & m_i = f(a)= \text{Sigmoid}(\lambda(\vc'_i- a)) = \frac{1}{1+e^{-\lambda(\vc'_i - a)}}.
    \label{eq:sig}
\end{align}

We add a hyperparameter \( \lambda \) to control the degree of approximation from Equation \ref{eq:sig} to a hard mask generation function. When \( \lambda \) tends to infinity, Equation (\ref{eq:sig}) approaches a hard mask generation function. We usually set \( \lambda \) to \( N \). Because when \( c'_i>a+3/N \) or \( c'_i<a-3/N \), \( |(m_i-\lfloor m_i \rceil)|<0.05 \). It means that except for the six elements whose importance values are around the pruning ratio, the masks of other elements are close to 0 or 1, where the approximation error is less than 0.05.
Therefore, \( \lambda=N \) is sufficient to approximate a hard mask generation function for our topk.

\begin{figure}[t]
    \begin{subfigure}{0.49\columnwidth}
        \includegraphics[width=\textwidth]{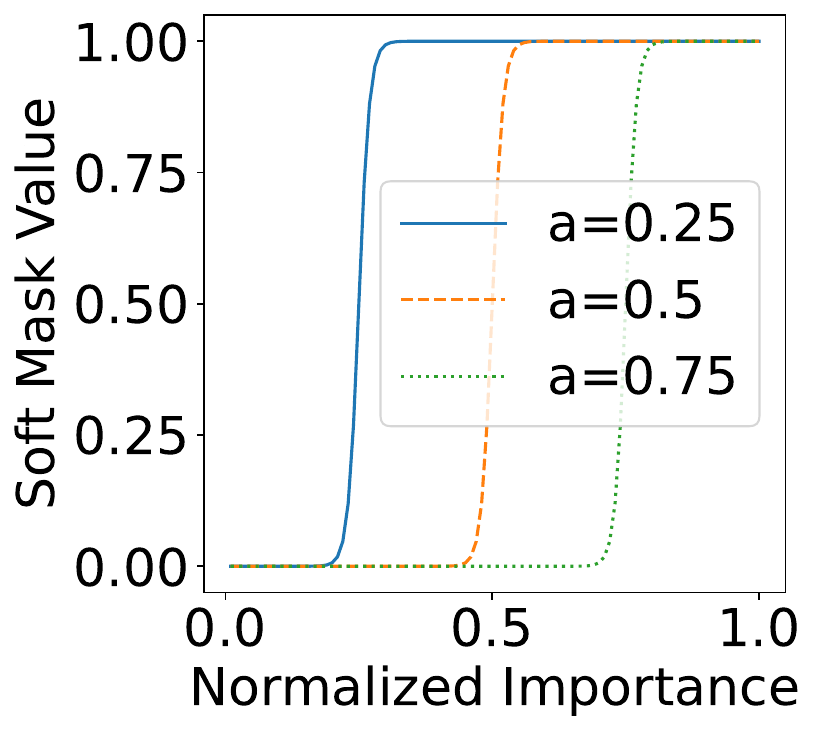}
        \caption{Forward}
    \end{subfigure}
    \begin{subfigure}{0.49\columnwidth}
        \includegraphics[width=\textwidth]{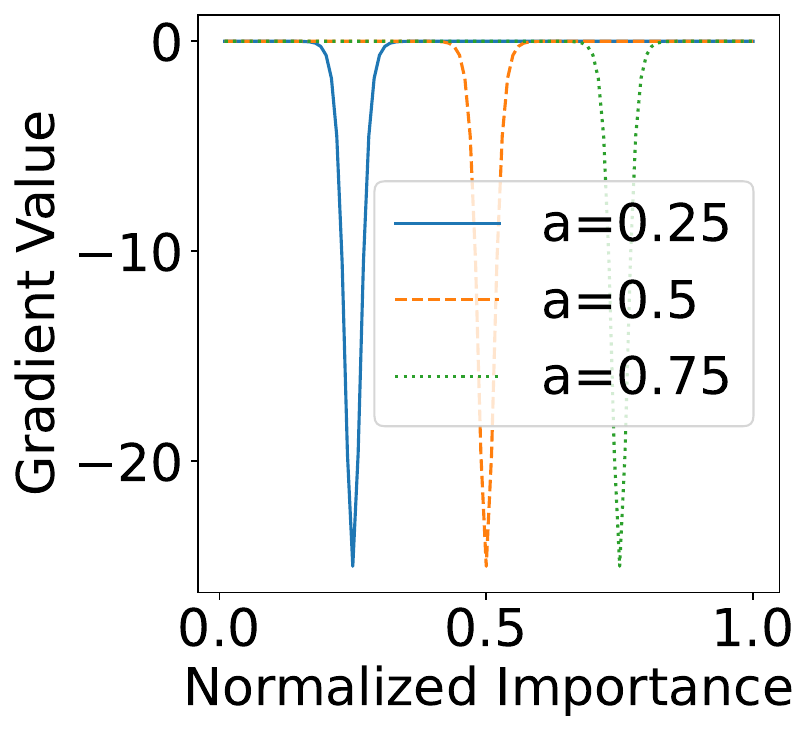}
        \caption{Backward}
    \end{subfigure}
    \caption{Forwad and Backward Graph of Our Differentiable Top-k. We set maximal element number \(N=\lambda=100 \), pruning ratio \(a\in \{0.25,0.5,0.75\} \).
        The x-axis represents the normalized element importance \(c'_i \).
        \textbf{(a)} demonstrates the forward process, where the y-axis represents the soft mask \(m_i \).
        \textbf{(b)} illustrates the backward process, where the y-axis represents the gradient of \(a \) with respect to \(m_i \), \(\frac{\partial m_i}{\partial a} \).}
    \label{fig:graph}
\end{figure}

The forward and backward graph of Equation \ref{eq:sig} are shown in Figure \ref{fig:graph} (a) and Figure \ref{fig:graph} (b), respectively.
It can be observed that
1) Our topk models the number of elements \(k\) directly using the learnable pruning ratio \(a\), and it generates a polarized soft mask \( \vm \) to simulate the pruned model perfectly during forward.
2) Our differentiable topk is fully differentiable and is able to be optimized stably. The gradient of \( a \) with respect to \( m_i \) is \(\frac{\partial m_i}{\partial a} = -\lambda(1-m_i)m_i \). Our topk intuitively detects the gradient of the mask in the fuzzy area with \( 0.05<m_i<0.95 \). Note, Figure \ref{fig:graph} (b) illustrates the value of \(\frac{\partial m_i}{\partial a} \), which
is not the total gradient of \(a\). The total gradient of \(a\) is
\( \sum_{i=1}^{N}{\frac{\partial task\_loss}{\partial m_i}\frac{\partial m_i}{\partial a}}+\frac{\partial resource\_loss}{\partial a}\).

\subsubsection{Element Evaluation}
As we do not limit the distribution of element importance, element importance can be quantified through various methods, such as L1-norm \cite{li2016l1}, among others. In our approach, we implement \emph{Taylor importance} \cite{molchanov2019tayler} in a \emph{moving average} manner as follows:

\begin{align}
    c^{t+1}_i = c^t_i \times decay + (m^t_i \times g_{i})^2 \times (1-decay).
\end{align}

Here, \( t \) represents the training step. \(g_i\) is the gradient of \(m_i\) with respect to training loss.
\( Decay \) refers to the decay rate. The initial value of \( c_i^0 \) is set to zero, and the decay rate is set to 0.99.
Note that the importance of elements is not updated by gradient descent.
By leveraging Taylor importance, we can efficiently and stably estimate the importance of elements.
We conduct an ablation study on the importance evaluation methods in \hyperref[sec:abla_metric]{Section 5.1}, which shows that Taylor importance is enough to achieve good performance.

\subsection{Differentiable Model Scaling}

Relying on our differentiable topk, we develop Differentiable Model Scaling (DMS) to optimize the width and depth of networks.
Our DMS has three pipeline variants based on that of training-based model pruning, as shown in Table \ref{tab:pipeline}.

\textbf{\DMSp} is the standard training-based model pruning pipeline, it consists of pretrain stage, search stage, and retrain stage.
The pretrain stage is used to pretrain a supernet. It usually costs a lot of time and resources.
In the search stage, we search for the optimal width and depth of the supernet under a specific resource constraint. Due to the high search efficiency of our method, the search stage only uses about 1/10 or fewer of the epochs of retraining.
In the retrain stage, we retrain the searched model.
We use this pipeline when comparing with SOTA pruning methods.

\textbf{\DMSnp} is our default and most-used pipeline in our paper.
The high costs of pretrain stage, which may make up most of the total costs, is a big obstacle to the practical application of NAS and pruning methods. To overcome this problem, we discard the pretrain stage from the \DMSp and directly search from a randomly initialized supernet.
According to our ablation study in \hyperref[sec:abla_structure]{Appendix A.3.2}, \DMSnp surpluses \DMSp by increasing the supernet size on both performance and efficiency. \DMSnp makes our method maintain high performance and is more efficient than other NAS methods.

\textbf{\DMSpn} is used to compare different search methods extremely quickly. Compared with \DMSp, it only optimizes the structural parameters and does not retrain the searched models. Take advantage of existing pretrained supernets, it also outputs reasonable results. Besides, it only takes hundreds of iterations, costing less than 10 minutes on a single RTX3090, to search for a model.

\begin{table}[t]
    \centering
    \caption{Three Pipelines We Used in Our Paper.}

    \vskip 0.1in
    \begin{small}
        \begin{tabular}{ l | c c  c}
            \toprule
            Pipeline & Pretrain & Search                     & Retrain \\
            \midrule
            \DMSnp   & N        & All weights                & Y       \\
            \DMSp    & Y        & All weights                & Y       \\
            \DMSpn   & Y        & Only structural parameters & N       \\
            \bottomrule
        \end{tabular}
    \end{small}
    \vskip -0.1in

    \label{tab:pipeline}
\end{table}

\textbf{Search Space}:
As shown in Figure \ref{fig:compare} (b), our search space encompasses both the width and depth of networks, which are the most critical structural hyperparameters for model scaling. To represent these dimensions, we use our differentiable topk.
The width in networks typically covers the channel dimension in convolutional layers, the feature dimension in fully connected layers, and so on.
Regarding depth, we focus on networks with residual connections and search the number of blocks in each stage.
Specifically, We incorporate the soft masks of differentiable topk into residual connections, allowing each block to be represented as \(x_{i+1}=x_i+f(x_i)\times m_i \).

Besides, for a structural hyperparameter $x$, we search it in the range of $[1, x_{max}]$ with a step of 1, while most of prior NAS methods \cite{cai2019once,chen2021autoformer} search it in the range of $[x_{min}, x_{max}]$ with a large step, like 32. Their search spaces have been a pretty good sub-space of ours, designed by human experts. However, our search space is more general and costs the least human effort.
We conduct an ablation study on search spaces in \hyperref[sec:abla_search_space]{Appendix A.3.6}. The results show that our method can achieve better performance on our fine-grained search space, which is harder to search, than prior methods on coarse-grained search space, which is easier to search.
More details about our search space are provided in \hyperref[sec:search_space]{Appendix A.1.1}.

\textbf{Resource Constraint Loss}
To ensure that a network adheres to specific \emph{resource constraints}, we incorporate an additional component into the optimization process, termed the ``\emph{resource constraint loss}''. Consequently, the aggregate loss function is:

\begin{align}
    loss= & loss_{task}+\lambda_{resource}\times loss_{resource}.                                                                                    \\
          & loss_{resource}=\begin{cases}
                                \log(\frac{r_{c}}{r_{t}}) & \text{if } r_{c} > r_{t} \\
                                0                         & \text{otherwise}
                            \end{cases}.
\end{align}

Here, \(loss_{task}\) denotes the \emph{task loss}. \(loss_{resource}\) represents the additional resource constraint loss, and the term \(\lambda_{resource}\) acts as its weighting factor.
\(r_c\) symbolizes the current level of resource consumption, which is calculated based on the learnable parameters of differentiable topk operators.
\(r_t\) denotes the targeted level of resource consumption and is user-specified.
As our topk is fully differentiable, the learnable structural parameters can be optimized under the guidance of both task loss and resource constraint loss.
More details about our resource constraint loss are provided in \hyperref[sec:constrait]{Appendix A.1.2}.
MAC constraint loss is used in most experiments.
We use latency constraint loss in \hyperref[sec:abla_latency]{Appendix A.3.3}, and use the number of parameters constraint loss in \hyperref[sec:llm]{Appendix A.2.4}. They show that our method is compatible with various resource constraints.

\section{Experiment}

We applied our method to rigorous evaluations across various tasks, including vision and NLP tasks, and architectures, including CNNs and Transformers.
We focus on search costs and performance.
The search cost associated with a model is divided into two distinct components: the \emph{public cost} that costs once for all searched models for a method, like supernet pretraining, and the \emph{private cost} that costs for each model.
Notably, our method consistently outperforms both baseline models and prior NAS methods, highlighting its superior performance and adaptability.

\subsection{Comparison with Different Search Methods}
\label{sec:ex_search}

To demonstrate the higher search efficiency of our method, we first compared it with different search methods in the same setting.
For simplicity and fairness, we utilize ``\DMSpn'' pipeline for all methods in this section, which loads pretrained weights. As all methods load pretrained weights, we don't count pretrain costs in search costs in this section.
All experiments in this section are conducted on ImageNet \cite{deng2009imagenet}.

\subsubsection{Comparison with Gradient-based Methods}
\label{sec:gradient_methods}

Firstly, we compare our method with other gradient-based methods, including the multiple element selection (MES), single number selection (SNS), and gradient estimate topk (GET). We use ResNet-50 as our supernet and search for models with about 3G MACs. The search stage costs 800 iterations for all methods.
The results are shown in Table \ref{tab:gradient}.

Our method achieves the best performance among all gradient-based methods by a large margin. Compared with the multiple element selection and single number selection, our method uses much fewer (less than 1/250) learnable parameters, making it easier to optimize. Compared with the gradient estimate topk, our parameters get more accurate gradients, achieving much better performance. These results demonstrate our ``direct and differentiable'' manner is much more efficient than other gradient-based methods.

\begin{table}[t]
    \centering
    \caption{Comparison with other Gradient-based Methods:
        ``MES'' means multiple element selection, ``SNS'' means single number selection, and ``GET'' means gradient estimate topk.
        "N" means the number of learnable structural parameters.
        We use ``\DMSpn'' pipeline for all methods in this table.
    }

    \vskip 0.1in
    \begin{small}
        \begin{tabular}{ l | c c | c}
            \toprule
            Method                    & Top-1 (\%)    & Macs (G) & N     \\
            \midrule
            \thead[l]{MES                                                \\ \cite{herrmann2020gumbel}}   & 55.5          & 3.2      & 11468 \\
            SNS \cite{wan2020fbnetv2} & 58.6          & 3.4      & 11468 \\
            GET \cite{yao2021leap}    & 61.6          & 3        & 41    \\
            \DMSpn (ours)             & \textbf{70.7} & 3        & 41    \\
            \bottomrule
        \end{tabular}
    \end{small}
    \vskip -0.1in

    \label{tab:gradient}
\end{table}

\subsubsection{Comparison with Evolutionary Algorithm}
\label{sec:evolution}

We additionally compare our method with an \emph{evolutionary algorithm (EA)}, a typical stochastic search method. We utilize the supernet of OFA \cite{cai2019once} and search for models with 0.45G MACs. The results are shown in Table \ref{tab:evolution}.
Our DMS just consumes less than 1/20 of the search cost of the evolutionary algorithm without any performance degradation.
It reveals that our method is more efficient than the evolutionary algorithm under the guidance of gradients.

\begin{table}[t]
    \centering
    \caption{Comparison with Evolutionary Algorithm:
        ``EA'' means evolutionary algorithm.
        Only EA + predictor needs a public search cost, which is used to train an accuracy predictor.
        The unit for search cost is GPU hours.
        We use the pipeline of ``\DMSpn'' in this table.
    }

    \vskip 0.1in
    \begin{small}
        \begin{tabular}{ l | c c | c}
            \toprule
            Method        & Top-1 (\%) & Macs (G) & \thead{Cost          \\ Public+Private}         \\
            \midrule
            \thead[l]{EA                                     + predictor \\ \cite{cai2019once}}  & 78.2       & 0.45     & 40 + 0            \\
            \thead[l]{EA                                                 \\ \cite{cai2019once}}  & 78.2       & 0.45     & 0 + 1.1           \\
            \DMSpn (ours) & 78.2       & 0.45     & \textbf{0 + 0.05}    \\
            \bottomrule
        \end{tabular}
    \end{small}
    \vskip -0.1in

    \label{tab:evolution}
\end{table}

\subsection{Comparison with SOTA NAS Methods}

\begin{table*}[t]
    \begin{center}
        \caption{Experiments on EfficientNet. We compare our DMS with other NAS methods on EfficientNet variants.
            We divide all NAS methods into two groups, including low-search-cost methods and high-search-cost methods.
            Our method outperforms low-search-cost methods by a large margin with similar search costs, and it uses much fewer search costs than high-search-cost methods achieving comparable or better performance.
            The unit of search cost is TPU days for EfficientNet and GPU days for other models.
            ``Ratio'' stands for the ratio of the search cost of the model to that of our corresponding DMS model.
            $\ddagger$ means the model is trained with distillation.
            How to obtain these search costs is detailed in \hyperref[sec:ap_cost]{Appendix A.4.2}.
            Note we use our default ``\DMSnp'' pipeline, which does not load pretrained supernets, in this table.
        }
        \vskip 0.1in
        \begin{small}

            \begin{tabular}{ l | c |  c c c r @{${}{}$} r r | c }
                \toprule
                Model                                              & NAS Type  & Top-1 (\%)    & MACs (G) & Params (M) & \multicolumn{2}{c}{\thead{ Search Cost                                                                           \\Public + Private}} & Ratio        & \thead{Cost \\ Level}                                       \\
                \midrule
                JointPruning$^\ddagger$ \cite{guo2021jointpruning} & Gradient  & 77.3          & 0.34     & /          & 0 +                                    & 8            & 2.5$\times$                     & \multirow{5}{*}{Low}   \\
                \textbf{\DMSnp-EN-350$^\ddagger$ (ours)}           & Gradient  & \textbf{78.5} & 0.35     & 5.6        & \textbf{0 +}                           & \textbf{3.2} & \textbf{1\boldsymbol{$\times$}}                          \\

                Zen-score$^\ddagger$ \cite{lin2021zen}             & ZeroShot  & 78.0          & 0.41     & 5.7        & 0 +                                    & 0.5          & 1.3$\times$                                              \\
                ZiCo$^\ddagger$ \cite{li2023zico}                  & ZeroShot  & 78.1          & 0.45     & /          & 0 +                                    & 0.4          & 1$\times$                                                \\
                \textbf{\DMSnp*-EN-450$^\ddagger$ (ours)}          & Gradient  & \textbf{79.4} & 0.45     & 6.5        & \textbf{0 +}                           & \textbf{0.4} & \textbf{1\boldsymbol{$\times$}}                          \\
                \hline
                ScaleNet-EN-B0 \cite{xie2022scalenet}              & OneShot   & 77.5          & 0.35     & 4.4        & 379 +                                  & 1.6          & 119$\times$                     & \multirow{11}{*}{High} \\
                \textbf{\DMSnp-EN-350 (ours)}                      & Gradient  & \textbf{78.0} & 0.35     & 5.6        & \textbf{0 +}                           & \textbf{3.2} & \textbf{1\boldsymbol{$\times$}}                          \\

                EfficientNet-B0 \cite{tan2019efficientnet}         & MultiShot & 77.1          & 0.39     & 5.3        & 1714 +                                 & 0            & 536$\times$                                              \\
                \textbf{\DMSnp-EN-B0 (ours)}                       & Gradient  & \textbf{78.5} & 0.39     & 6.2        & \textbf{0 +}                           & \textbf{3.2} & \textbf{1\boldsymbol{$\times$}}                          \\

                EfficientNet-B1 \cite{tan2019efficientnet}         & MultiShot & 79.1          & 0.69     & 7.8        & 1714 +                                 & 0            & 296$\times$                                              \\
                ScaleNet-EN-B1 \cite{xie2022scalenet}              & OneShot   & 79.9          & 0.80     & 7.4        & 379 +                                  & 3.7          & 66$\times$                                               \\
                MA-EN-B1 \cite{liu2022modelamplification}          & MultiShot & 79.9          & 0.68     & 8.8        & $>$ 124 +                              & \;131        & $>$ 44$\times$                                           \\
                \textbf{\DMSnp-EN-B1 (ours)}                       & Gradient  & \textbf{80.0} & 0.68     & 8.9        & \textbf{0 +}                           & \textbf{5.8} & \textbf{1\boldsymbol{$\times$}}                          \\

                EfficientNet-B2 \cite{tan2019efficientnet}         & MultiShot & 80.1          & 1.0      & 9.2        & 1714 +                                 & 0            & 245$\times$                                              \\
                MA-EN-B2 \cite{liu2022modelamplification}          & MultiShot & 80.9          & 1.0      & 9.3        & $>$ 124 +                              & 192          & $>$ 45$\times$                                           \\
                \textbf{\DMSnp-EN-B2 (ours)}                       & Gradient  & \textbf{81.1} & 1.1      & 9.6        & \textbf{0 +}                           & \textbf{7.0} & \textbf{1\boldsymbol{$\times$}}                          \\
                \bottomrule
            \end{tabular}

        \end{small}
        \vskip -0.1in
        \label{tab:eff}
    \end{center}
\end{table*}

Then, we compare our method with SOTA NAS methods on ImageNet.
We choose EfficientNet models as baselines and search for optimal configurations in terms of their width and depth. EfficientNet \cite{tan2019efficientnet} is a widely accepted baseline for NAS research \cite{xie2022scalenet,liu2022modelamplification}.
We use our default pipeline ``\DMSnp'', which does not load pretrained weights, in this section, Therefore, the public search cost for our method is zero.

The performance of these searched models and their search costs are presented in Table \ref{tab:eff}.
According to the level of their search costs, we divide all compared NAS methods into two groups: \emph{low-search-cost methods} and \emph{high-search-cost methods}.
Low-search-cost methods usually contain gradient-based and zero-shot NAS methods. This category only needs several GPU days, which are fewer than that of training the searched model itself.
high-search-cost methods usually contain multi-shot and one-shot NAS methods, which usually cost more than one hundred GPU days (much over than training the searched model itself).

\textbf{Compared with Low-Search-Cost Methods}:
Our method also belongs to low-search-cost methods. Compared with this category, our method achieves better performance with similar or even lower search costs. For example, our method outperforms JointPruning \cite{guo2021jointpruning} by 1.2\% with 2/5 of its search cost. It also outperforms zero-shot NAS methods, ZiCo and Zen-score, by a margin of 1.3\% and 1.4\%, respectively. This result demonstrates that our method achieves significant performance improvements than previous low-search-cost NAS methods.

\textbf{Compared with High-Search-Cost Methods}:
Compared with this category, our method spends much fewer search costs but still achieves comparable or even better performance. For example, Compared with EfficientNet, our searched models, \DMSnp-EN-B0, B1, and B2, have improved performance by 1.4\%, 0.9\%, and 1.0\%, respectively. Remarkably, DMS also achieves over 100 times search cost savings in the search process. our method outperforms ScaleNet by 0.5\% and 0.1\% on EfficientNet-B0 and B1, respectively, with less than 1/50 of the search cost of it.
Our DMS also outperforms ModelAmplification (MA) by 0.1\% and 0.2\% on EfficientNet-B1 and B2, respectively, with less than 1/40 of the search cost of it.
This result proves the high search efficiency of our method.

For high-search-cost methods, their search costs usually come from huge public costs, like supernet training. Even if we average their public cost over four variants (the number of variants used by most papers), they still cost over 20 times our total search cost. Except for the drawback of high search costs, the high-public-cost methods make it impossible to search a large model. For example, if a target model needs a month to train from scratch, like a LLM, the public search process of high-search-cost methods may take over a year, which is unacceptable for practical applications. While our method only needs a few days to search for a model, making it possible to search for a large model.

Through these experiments, we demonstrate that our method is more suitable for real-world applications,
as it achieves comparable or even better performance than SOTA NAS methods with low search costs.
It makes it easy to be inserted into the practical pipeline of model development.
We compare our method with more NAS methods in \hyperref[sec:more]{Appendix A.2.1} and draw this table as accuracy vs MACs plots and a search cost vs accuracy plot in Figure \ref{fig:macs}.

\subsection{Comparison with SOTA pruning methods}
\label{sec:pruning}

\begin{table}
    \centering
    \caption{Comparison with SOTA Pruning Methods. The supernet is ResNet-50 for all methods. We use ``\DMSp'' pipeline, which loads pretrained weights as other methods, in this table.}

    \vskip 0.1in
    \begin{small}
        \begin{tabular}{ l | c c}
            \toprule
            Method              & MACs (G) & Top-1 (\%)     \\
            \midrule
            ResNet-50           & 4.1      & 76.5           \\
            \midrule
            LFPC \cite{lfpc}    & 1.6      & 74.46          \\
            GReg\-2 \cite{greg} & 1.6      & 74.93          \\
            CC \cite{cc}        & 1.5      & 74.54          \\
            TPP \cite{tpp}      & 1.6      & 75.12          \\
            \DMSp(ours)         & 1.6      & \textbf{75.53} \\
            \bottomrule
        \end{tabular}

    \end{small}
    \vskip -0.1in
    \label{tab:pruning}
\end{table}

Our method can also be applied as a model structure pruning method.
We compare our method with SOTA pruning methods in Table \ref{tab:pruning}.
Following prior structure pruning methods, we utilize  ``\DMSp'' pipeline, which loads pretrained weights, and only prune width.

Compared with SOTA pruning methods, our DMS achieves the best performance though we do not employ complicated importance evaluation methods like others.
This is because of the strong search ability of our differentiable topk.

\textbf{More experiments}:
Except for the above experiments, we also applied our method to more architectures and more tasks. The architectures include ResNet \cite{he2016resnet}, MobileNetV2 \cite{sandler2018mobilenetv2} and Deit \cite{deit} (detailed in \hyperref[sec:archs]{Appendix A.2.2}). The tasks include object detection (detailed in \hyperref[sec:detection]{Appendix A.2.3}) and language modeling (detailed in \hyperref[sec:llm]{Appendix A.2.4}). Our method improves yolov8-n by 2.0\% on COCO and improves pruned Llama-7B on several language modeling benchmarks.
These results show that our method is highly versatile, rather than just being a method for one task or one architecture.

\section{Ablation Study}

\subsection{Ablation Study on Element Importance Metrics}
\label{sec:abla_metric}

Here, we compare different element importance metrics, including Index metric, SNIP \cite{lee2018snip}, Fisher \cite{liu2021group}, and Taylor \cite{molchanov2019tayler}.
We use the pipeline of ``\DMSnp'', which does not load pretrained weights, in this section.
We search for 1G models on ResNet-50 and compare the performance after retraining.

Index metric is the simplest metric, which assigns importance value according to the index of elements statically. SNIP, Fisher, and Taylor are more complicated metrics, using gradients and activations to update the importance of elements with a moving average.
The results are shown in Table \ref{tab:metric}. Index metric works poorly as its static strategy. SNIP, Fisher, and Taylor work better than the index metric and are comparable with each other for our method.

We further compare Taylor importance with and without moving average. The results are shown in Table \ref{tab:metric}. It can be seen that moving average can improve the performance of Taylor importance, because the moving average can smooth the importance of elements, making it more stable.

Therefore we just apply Taylor importance as our default metric, as it's more widely used in prior works \cite{humble2022soft,molchanov2019tayler}.

\begin{table}[t]
    \caption{Ablation Study on Element Importance Metric. We use the pipeline of ``\DMSnp'' in this table.}
    \begin{center}

        \vskip 0.1in
        \begin{small}
            \begin{tabular}{ l | c }
                \toprule
                Element Importance                & Top-1 (\%)    \\
                \midrule
                Index metric                      & 72.3          \\
                \midrule
                SNIP \cite{lee2018snip}           & 73.0          \\
                Fisher \cite{liu2021group}        & \textbf{73.2} \\
                \midrule
                Taylor w/o moving average         & 72.5          \\
                Taylor \cite{molchanov2019tayler} & 73.1          \\
                \bottomrule
            \end{tabular}
        \end{small}
        \vskip -0.1in

    \end{center}

    \label{tab:metric}
\end{table}

\textbf{More Ablation}: There are more ablation studies in the \hyperref[sec:ablation_more]{Appendix A.3}:
(1) Ablation study on importance normalization, in \hyperref[sec:abla_normal]{Appendix A.3.1}, demonstrate that importance normalization is significant for our method.
(2) Ablation study on pretraining and supernet sizes, in \hyperref[sec:abla_structure]{Appendix A.3.2}, shows that our \DMSnp, which does not load pretrained weights, outperforms \DMSp, which loads pretrained weights, on efficiency and efficacy by increasing supernet size.
(3) Experiments with latency constraint, in \hyperref[sec:abla_latency]{Appendix A.3.3}, shows that our method is also compatible with latency constraints.
(4) Ablation study on search time, in \hyperref[sec:abla_search_time]{Appendix A.3.4}, shows that our method is efficient in terms of search time.
(5) Ablation study on our hyperparameters, in \hyperref[sec:abla_hyper]{Appendix A.3.5}, shows that our method has few hyperparameters that need to be tuned for each model, and they are easy to set.
(6) Ablation study on search spaces, in \hyperref[sec:abla_search_space]{Appendix A.3.6}, shows that our method is compatible with other search spaces. Besides, our method can achieve better performance on our fine-grained search space, which is harder to search, than prior methods on coarse-grained search space.

\section{Conclusion}

In this paper, we introduce a novel model scaling method termed \emph{Differentiable Model Scaling (DMS)}.
Compared with prior NAS methods, our DMS has three advantages.
(1) DMS is efficient for searching, which makes it easy to use.
(2) DMS also achieves high performance, comparable with SOTA NAS methods.
(3) DMS is universal and is compatible with various tasks and architectures.
In conclusion, our DMS is a highly efficient and versatile method for model scaling, which is suitable for real-world applications.

\section*{Acknowledgements}

We express our appreciation to our colleague Zhihao Lin for his insightful discussions. We are grateful to the anonymous reviewers for their constructive feedback and recommendations.

\section*{Impact Statement}

This paper presents work whose goal is to advance the field
of Machine Learning. There are many potential societal
consequences of our work, none which we feel must be
specifically highlighted here.

\bibliography{example_paper}
\bibliographystyle{icml2024}

\newpage
\appendix
\onecolumn

\section{Appendix}

\subsection{More Details about DMS}

\label{sec:dms}

\subsubsection{Search Space}

\label{sec:search_space}

Our search space encompasses both the width and depth of networks, which are the most critical structural hyperparameters for model scaling.

The width in networks typically covers the channel dimension in convolutional layers, the feature dimension in fully connected layers, qkv dimension and the number of heads in attention mechanisms, among others. For convolutional and fully connected layers, we use two distinct differentiable topk operators to model their respective input and output widths, treating each channel or feature as an individual element. For multi-head attention, we employ a single differentiable topk to represent the number of heads, treating each head as a separate element.

Specifically, We apply our differentiable topk to different layers by multiplying masks, output by differentiable topk operators, with inputs to layers.
For convolutional layers, suppose the input is $X \in \mathbb{R}^{B \times C \times H \times W}$, and the mask is reshaped as $m \in \mathbb{R}^{1 \times C \times 1 \times 1}$, $X \times m$ works as the new input to the layer.
For an attention layer, we search the head dims of qkv and the number of heads. Suppose our supernet has $H$ heads and $D$ dims in each head. We have a mask for qk head dim with $m_{qk}\in R^{1\times 1 \times 1 \times D}$, a mask for v head dim with $m_{v}\in R^{1\times 1 \times 1 \times D}$, and a mask for number of heads $m_{head}\in R^{1\times H \times 1 \times 1}$. Suppose the sequence length is $L$, and the qkv for self-attention is $Q,K,V \in R^{B \times H \times L \times D}$. We compute the output of the self-attention by $softmax(\frac{Q'K'^T}{\sqrt{D}})V'$, where $Q'=Q\times m_{qk}\times m_{head}, K'=K\times m_{qk}\times m_{head}, V'=V\times m_{v}\times m_{head}$.

It is crucial to highlight that there can be channel or feature dependencies within models \cite{liu2021group,fang2023depgraph}. Interdependent Layers are treated as one group and share the same differentiable topk. We implemented this using an open-source model compression toolkit MMRazor \cite{2021mmrazor}, which is able to build element dependencies automatically.

Regarding depth, we focus on networks with residual connections. In this context, a residual block can be defined as \(x_{i+1}=x_i+f(x_i) \), and contiguous residual blocks are viewed as a network stage. The depth in our approach mainly comprises the number of blocks in each stage. We use a single differentiable topk for a network stage, with each block functioning as a distinct element. We incorporate the soft masks of differentiable topk into residual connections, allowing each block to be represented as \(x_{i+1}=x_i+f(x_i)\times m_i \).
In the context of Transformers, an attention mechanism combined with a feed-forward network (FFN) is considered as one block sharing the same soft mask.

The depth and width structure hyperparameters are trained jointly in our approach. For example, we have a $layer$ and an input $x$; we use $m_{L_i} \in [0,1]$ to denote the depth mask and $m_{C}\in [0,1]^{N}$ for the width mask.
The forward process is as follows: $y= m_C \times x+m_{L_i} \times layer(m_C \times x)$.
After searching, we will prune depth and width according to the depth mask and width mask, respectively.

Besides, as some maximal numbers of elements are small from several to tens, like the number of blocks and attention heads, we increase the \(\lambda\) in the differentiable topk operators from \(N\) to \(4N\) to approximate a hard mask generation function better.

For a structural hyperparameter $x$, we search it, in a fine-grained manner, in the range of $[1,x_{max}]$ with a step of 1. while most of prior NAS methods \cite{cai2019once,chen2021autoformer} search it, in a coarse-grained manner, in the range of $[x_{min},x_{max}]$ with a large step, like 32.
Our search method is also comparable with these coarse-grained search spaces.
We limit the value of $a$ to $[0,1-\frac{x_{min}}{x_{max}}]$ to ensure $x\in [x_{min}, x_{max}]$.
We treat contiguous elements in a step as an unit, and each unit share the same element importance and mask. For example, when searching the number of channels in a layer with a step of 32, each 32 channels share the same element importance and mask.

\subsubsection{Resource Constrait}

\label{sec:constrait}

Our resource constraint loss is defined as:
\begin{align}
    loss_{resource}= & \begin{cases}
                           \log(\frac{r_{c}}{r_{t}}) & \text{if } r_{c} > r_{t} \\
                           0                         & \text{otherwise}
                       \end{cases}. \\
\end{align}

In this definition, \(r_c\) symbolizes the current level of resource consumption, and \(r_t\) denotes the targeted level of resource consumption. If \(r_c\) exceeds \(r_t\), a non-zero \(loss_{resource}\) is used to compress the model.
The value of \(r_t\) is user-specified.
The value of \(r_c\) is calculated based on the learnable parameters of differentiable topk operators.
Take a linear layer as an example, we use \(f_{in}\) and \(f_{out}\) to represent the number of input and output features, respectively. \(a_{in}\) and \(a_{out}\) are the learnable parameters of differentiable topk operators for that layer.
We compute different resource consumption as follows:

\textbf{MAC constraint:} \(r_c=f_{in}\times a_{in} \times f_{out} \times a_{out}\times batchsize\)

\textbf{The Number of Parameters constraint:} \(r_c=f_{in}\times a_{in} \times f_{out} \times a_{out}\)

\textbf{Latency constraint:} \(r_c=latency_{max}\times F(a_{in},a_{out})\), where $latency_{max}$ is the latency of the layer without pruning. $F$ is a contiguous function to map the pruning ratio \(a_{in}\),\(a_{out}\) to the ratio of the latency of the pruned layer to the original layer. $F$ can be found by:
\begin{align}
    F=argmin_{F}(MSE(latency_{pruned(a_{in},a_{out})}-latency_{max}\times F(a_{in},a_{out})))
\end{align}
$latency_{pruned(a_{in},a_{out})}$ is the latency of the pruned layer with \(a_{in}\),\(a_{out}\) as the pruning ratio.
Take the OFA search space as an example, \(a_{out}\) always equals \(a_{in}\), and each block only has three width choices. Hence, we just employ a quadratic function as \(F\), and it works fine.

To enhance stability during training, we gradually reduce \(r_t\) to the final target value \(r_t^{final}\) throughout the training process by default.
For an epoch-based training procedure, \(r_t\) is determined by an exponential decay function as shown below:

\begin{align}
    {r_{t}}=(\frac{r_{t}^{final}}{r_{supernet}})^{\frac{e}{e_{max}}} \times r_{supernet},\quad \text{where } \frac{r_{t}^{final}}{r_{supernet}}<1.
\end{align}

Here, \(e\) denotes the current epoch out of total epochs \(e_{max}\). \(r_{supernet}\) is a constant representing the resource demand of the supernet.

Furthermore, since resource consumption can fluctuate significantly with respect to depth, we introduce extra epochs dedicated to optimizing width while maintaining depth constant.
By adopting the strategies outlined above, Differentiable Model Scaling ensures that models adhere to specific resource constraints.

\subsection{More Experiments}

\subsubsection{Comparison with More NAS Methods}

\label{sec:more}

We additionally compare our method with more NAS methods, as shown in Table \ref{tab:eff_more}. Note this is a rough comparison. As it's hard to compute a precise search cost for some methods, we only group them into two groups, including high-search-cost methods and low-search-cost methods, according to a rough estimation of their search costs. High-search-cost methods costs over 100 GPU days. while others belong to low-search-cost methods.

\begin{figure}
    \begin{center}
        \begin{subfigure}{0.33\columnwidth}
            \includegraphics[width=\textwidth]{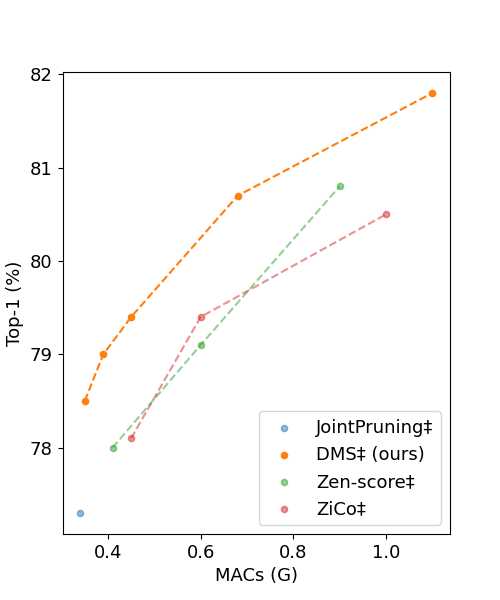}
            \subcaption{}
        \end{subfigure}
        \begin{subfigure}{0.33\columnwidth}
            \includegraphics[width=\textwidth]{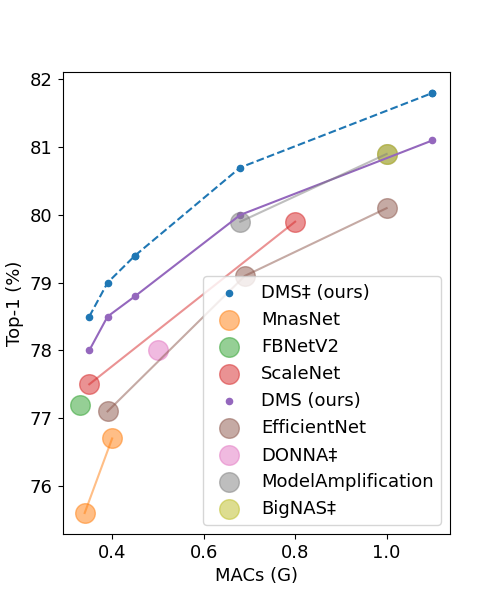}
            \subcaption{}
        \end{subfigure}
        \begin{subfigure}{0.33\columnwidth}
            \includegraphics[width=\textwidth]{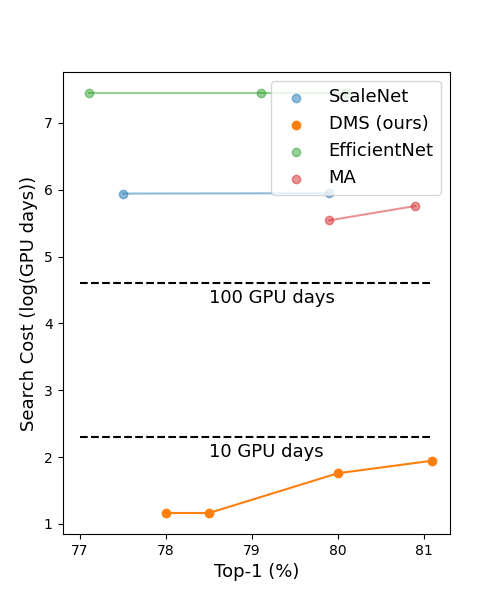}
            \subcaption{}
        \end{subfigure}

        \caption{
            We draw these three plots based on Table \ref{tab:eff} and Table \ref{tab:eff_more}. We use larger dot sizes to represent the "High" search cost level and smaller dot sizes to represent the "Low" search cost level. Dashed lines are used to represent the models trained with distillation.
            \textbf{(a) Performance Comparison with Low-Search-Cost methods} It can be seen that our method outperforms these methods significantly.
            \textbf{(b) Performance Comparison with High-Search-Cost methods} Our method achieves comparable or even better performance, while all high-search-cost methods cost more than dozens of times our total search costs.
            \textbf{(c) Search Cost Comparison with High-Search-Cost methods} We compare the search costs of ours and that of high-search-cost methods. We only draw methods with precise search cost estimation. The search costs of unpainted high-search-cost methods are also larger than 100 GPU days. We present search costs on a log scale.
        }
        \label{fig:macs}
    \end{center}
\end{figure}

We draw Table \ref{tab:eff} and Table \ref{tab:eff_more} as accuracy vs MACs plots and a search costs vs accuracy plot, as shown in Figure \ref{fig:macs}.
It can be observed that our DMS outperforms low-search-cost methods significantly. DMS achieves much lower search costs than high-search-cost methods, achieving comparable and even higher performance.

\begin{table}
    \caption{Rough Comparison with more NAS methods. As some methods did not report their search cost, we simply use "High" and "Low" to represent the search cost of NAS methods, while "High" for multi-shot NAS methods and one-shot NAS methods, "Low" for gradient-based NAS methods and zero-shot NAS methods. $\ddagger$ means the model is trained with distillation. We use the pipeline of ``\DMSnp'', which does not load pretrained weights, in this table}

    \begin{center}
        \vskip 0.1in
        \begin{small}
            \begin{tabular}{ l | c |  c  c c | c }
                \toprule
                Model                                                     & NAS Type  & Top-1 (\%)    & MACs (G) & Params (M) & Cost Level             \\
                \midrule
                JointPruning$^\ddagger$  \cite{guo2021jointpruning}       & Gradient  & 77.3          & 0.34     & /          & \multirow{13}{*}{Low}  \\
                \textbf{\DMSnp-EN-350$^\ddagger$ (ours)}                  & Gradient  & \textbf{78.5} & 0.35     & 5.6        &                        \\
                Zen-score$^\ddagger$ \cite{lin2021zen}                    & ZeroShot  & 78.0          & 0.41     & 5.7        &                        \\
                \textbf{\DMSnp-EN-B0$^\ddagger$ (ours)}                   & Gradient  & \textbf{79.0} & 0.39     & 6.2        &                        \\
                ZiCo$^\ddagger$ \cite{li2023zico}                         & ZeroShot  & 78.1          & 0.45     & /          &                        \\
                \textbf{\DMSnp*-EN-450$^\ddagger$ (ours)}                 & Gradient  & \textbf{79.4} & 0.45     & 6.5        &                        \\
                Zen-score$^\ddagger$ \cite{lin2021zen}                    & ZeroShot  & 79.1          & 0.60     & 7.1        &                        \\
                ZiCo$^\ddagger$ \cite{li2023zico}                         & ZeroShot  & 79.4          & 0.60     & /          &                        \\
                \textbf{\DMSnp-EN-B1$^\ddagger$ (ours)}                   & Gradient  & \textbf{80.7} & 0.68     & 8.9        &                        \\
                ZiCo$^\ddagger$ \cite{li2023zico}                         & ZeroShot  & 80.5          & 1.0      & /          &                        \\
                Zen-score$^\ddagger$ \cite{lin2021zen}                    & ZeroShot  & 80.8          & 0.9      & 19.4       &                        \\
                \textbf{\DMSnp-EN-B2$^\ddagger$ (ours)}                   & Gradient  & \textbf{81.8} & 1.1      & 9.6        &                        \\
                \midrule
                MnasNet-A2 \cite{tan2019mnasnet}                          & MultiShot & 75.6          & 0.34     & 4.8        & \multirow{22}{*}{High} \\
                FBNetV2-L1 \cite{wan2020fbnetv2}                          & Gradient  & 77.2          & 0.33     & /          &                        \\
                ScaleNet-EN-B0 \cite{xie2022scalenet}                     & OneShot   & 77.5          & 0.35     & 4.4        &                        \\
                \textbf{\DMSnp-EN-350 (ours)}                             & Gradient  & \textbf{78.0} & 0.35     & 5.6        &                        \\
                MnasNet-A3 \cite{tan2019mnasnet}                          & MultiShot & 76.7          & 0.40     & 5.2        &                        \\
                EfficientNet-B0 \cite{tan2019efficientnet}                & MultiShot & 77.1          & 0.39     & 5.3        &                        \\
                \textbf{\DMSnp-EN-B0 (ours)}                              & Gradient  & \textbf{78.5} & 0.39     & 6.2        &                        \\
                DONNA$^\ddagger$ \cite{moons2021donna}                    & OneShot   & 78.0          & 0.50     & /          &                        \\
                \textbf{\DMSnp*-EN-450 (ours)}                            & Gradient  & \textbf{78.8} & 0.45     & 6.5        &                        \\
                \textbf{\DMSnp*-EN-450$^\ddagger$ (ours)}                 & Gradient  & \textbf{79.4} & 0.45     & 6.5        &                        \\
                EfficientNet-B1 \cite{tan2019efficientnet}                & MultiShot & 79.1          & 0.69     & 7.8        &                        \\
                \textbf{\DMSnp-EN-B1 (ours)}                              & Gradient  & \textbf{80.0} & 0.68     & 8.9        &                        \\
                ScaleNet-EN-B1 \cite{xie2022scalenet}                     & OneShot   & 79.9          & 0.80     & 7.4        &                        \\
                ModelAmplification-EN-B1 \cite{liu2022modelamplification} & MultiShot & 79.9          & 0.68     & 8.8        &                        \\
                EfficientNet-B2 \cite{tan2019efficientnet}                & MultiShot & 80.1          & 1.0      & 9.2        &                        \\
                ModelAmplification-EN-B2 \cite{liu2022modelamplification} & MultiShot & 80.9          & 1.0      & 9.3        &                        \\
                BigNAS-XL$^\ddagger$ \cite{liu2022modelamplification}     & OneShot   & 80.9          & 1.0      & 9.5        &                        \\
                \textbf{\DMSnp-EN-B2 (ours)}                              & Gradient  & \textbf{81.1} & 1.1      & 9.6        &                        \\
                \textbf{\DMSnp-EN-B2$^\ddagger$ (ours)}                   & Gradient  & \textbf{81.8} & 1.1      & 9.6        &                        \\

                \bottomrule
            \end{tabular}
        \end{small}
        \vskip -0.1in

        \label{tab:eff_more}
    \end{center}
\end{table}

\subsubsection{Image Classification Experiments on More Architectures}
\label{sec:archs}

\begin{table}
    \caption{Experiments on ImageNet with Various Architectures. We searched the models' width and depth and compared
        them with the original models. We use the pipeline of ``\DMSnp'', which does not load pretrained weights, in this table.}

    \begin{center}
        \vskip 0.1in
        \begin{small}
            \begin{tabular}{ l | c  c  c }
                \toprule
                Model                                     & Top-1 (\%)    & MACs (G) & Params (M) \\
                \midrule
                ResNet-50 \cite{he2016resnet}             & 76.5          & 4.1      & 25.6       \\
                DMS-ResNet                                & \textbf{77.6} & 4.0      & 28.4       \\
                \midrule
                ResNet-50-rsb-a1 \cite{he2016resnet}      & 80.1          & 4.1      & 25.6       \\
                DMS-ResNet-rsb-a1                         & \textbf{81.0} & 4.0      & 28.4       \\
                \midrule
                MobileNetV2 \cite{sandler2018mobilenetv2} & 72.0          & 0.3      & 3.4        \\
                DMS-MobileNetV2                           & \textbf{73.0} & 0.3      & 5.3        \\
                \midrule
                Deit-T \cite{deit}                        & 74.5          & 1.3      & 5.7        \\
                DMS-Deit-T                                & \textbf{75.1} & 1.3      & 6.2        \\
                \bottomrule
            \end{tabular}
        \end{small}
        \vskip -0.1in
        \label{tab:models}
    \end{center}
\end{table}

To validate the universality of our method across various model architectures, we applied it to different architectures, as shown in Table \ref{tab:models}.

\textbf{Classic CNNs}: We validated our method on ResNet \cite{he2016resnet} and MobileNetV2 \cite{sandler2018mobilenetv2}. Our searched ResNet surpasses ResNet-50 by 1.1\%. Furthermore, when the searched ResNet is trained using an enhanced training setting (referred to as rsb-a1 \cite{wightman2021rsb}), it also exceeds the corresponding model by 0.9\%. Although MobileNetV2 is a lightweight model, our searched version outperforms the original model by a margin of 1.0\%.

\textbf{Transformer}: We additionally applied our method to Deit, a one-stage transformer.
Our searched models outperform the original model by 0.6\%.

These results demonstrate the strong ability of our DMS to search for the optimal structure of models.

\subsubsection{Object Detection Experiments on COCO}
\label{sec:detection}

\begin{table}
    \centering
    \caption{Object Detection Experiments on COCO. We use the pipeline of ``\DMSnp'', which does not load pretrained weights, in this table.}
    \vskip 0.1in
    \begin{small}
        \begin{tabular}{ l | c c c}
            \toprule
            Model                   & mAP (\%)      & MACs (G) & Params (M) \\
            \midrule
            Yolo-v8-n \cite{yolov8} & 37.4          & 4.4      & 3.2        \\
            \DMSnp-Yolo-v8-n (ours) & \textbf{39.4} & 4.2      & 2.7        \\
            \bottomrule
        \end{tabular}
    \end{small}
    \vskip -0.1in
    \label{tab:detection}
\end{table}

Since the complete end-to-end searching of our differentiable topk, DMS is a general search method that can be applied to various tasks. We also evaluated DMS for object detection on COCO. We chose Yolo-v8-n \cite{yolov8} as the baseline model and searched for the optimal structure of it. Our searched version betters the original model by 2.0\% in box AP, as shown in Table \ref{tab:detection}.

\subsubsection{LLM Experiment}
\label{sec:llm}

\begin{table*}
    \caption{Experiment on Llama-7B. We pruned Llama-7B using DMS and compared it with LLMPruner.
        We evaluate the pruned model using perplexity on Wikitext2 and Pth datasets and zero-shot classification accuracy on BoolQ, WinoGrande, ARC-e, and ARC-c datasets.
        In our results, the symbol ``$\uparrow$'' denotes that a larger value is better, while ``$\downarrow$'' signifies that a smaller value is preferable.
        We use our pipeline of ``\DMSp'', which loads pretrained weights, in this table, due to resource constraints.}

    \begin{center}
        \vskip 0.1in
        \begin{small}
            \begin{tabular}{l | c | c  c|  c c c c}
                \toprule
                Model                                      & Params & Wikitext2 \(\downarrow \) & Pth \(\downarrow \) & BoolQ \(\uparrow \) & WinoGrande \(\uparrow \) & ARC-e \(\uparrow \) & ARC-c \(\uparrow \) \\
                \midrule
                Llama-7B \cite{touvron2023llama}           & 6.74B  & 12.62                     & 22.14               & 76.5                & 67.01                    & 72.8                & 41.38               \\
                \midrule
                LLM-Pruner-Llama-7B \cite{ma2023llmpruner} & 5.47B  & 17.39                     & 30.2                & 66.79               & 64.96                    & 64.06               & 37.88               \\
                \DMSp-Llama-7B (ours)                      & 5.47B  & \textbf{17.13}            & \textbf{27.98}      & \textbf{75.23}      & \textbf{65.35}           & \textbf{71.46}      & \textbf{39.59}      \\
                \bottomrule
            \end{tabular}
        \end{small}
        \vskip -0.1in

        \label{tab:llm}
    \end{center}
\end{table*}

Beyond vision tasks, we extended our method to evaluate its applicability on a large language model (LLM) called Llama \cite{touvron2023llama}, as shown in Table \ref{tab:llm}. Due to resource constraints, we were unable to train an LLM from scratch. Instead, we adopted a ``prune and finetune'' strategy using the alpaca dataset \cite{alpaca}. This is similar to our pipeline of ``\DMSp'', but only finetuning rather than retraining. To mitigate overfitting to the alpaca dataset, we used the original model to distill the pruned model both during pruning and the subsequent finetuning process.
In alignment with LLMPruner \cite{ma2023llmpruner}, we limited our pruning to the heads of self-attentions and the hidden dimensions of the feed-forward networks (FFN) within Llama. The resource constraint is set to 5.47B parameters as LLMPruner.
After pruning 20\% of the parameters from Llama-7B and comparing it with LLMPruner, our method demonstrated superior performance across various benchmarks. Specifically, we observed reduced perplexity on WikiText2 \cite{merity2016wiki} and Pth \cite{marcus1993ptb}, and higher zero-shot classification accuracy on BoolQ \cite{clark2019boolq}, WinoGrande \cite{sakaguchi2021winogrande}, ARC-e \cite{clark2018ace}, and ARC-c \cite{clark2018ace}.

\subsection{More Ablaion Study}
\label{sec:ablation_more}

\subsubsection{Ablation Study on Importance Normalization}
\label{sec:abla_normal}
We compare our method with that of disabling importance normalization (replace Eq \ref{eq:normalize} by \(c'_i=c_i\)), as shown in Table \ref{tab:norm}.
We use the pipeline of ``\DMSpn'' for all methods in this table, and ResNet-50 as our supernet.
Without importance normalization, our method drops to an extremely low performance. It's caused by the unevenly distributed importance, which makes our topk function hard to optimize.
This result demonstrates the significance of our importance normalization.

\begin{table}[t]
    \centering
    \caption{Comparison with other Gradient-based Methods:
        We use ``\DMSpn'' pipeline for all methods in this table.
    }

    \vskip 0.1in
    \begin{small}
        \begin{tabular}{ l | c c  c}
            \toprule
            Method                   & Top-1 (\%)    & MACs (G) \\
            \midrule
            \DMSpn w/o normalization & 0.1           & 3        \\
            \DMSpn (ours)            & \textbf{70.7} & 3        \\
            \bottomrule
        \end{tabular}
    \end{small}
    \vskip -0.1in

    \label{tab:norm}
\end{table}

\subsubsection{Ablation Study on Pretraining and Supernet Sizes}
\label{sec:abla_structure}

\begin{table}[t]
    \centering
    \caption{Ablation Study on Pretraining and Supernet Size. We search for 1G models in these experiments. $Cost_{pretrain}$ is the cost of pretraining a supernet, $Cost_{search}$ is the cost of searching a model. The unit of cost is $G MACs \times epochs$.
    }
    \vskip 0.1in
    \begin{small}
        \begin{tabular}{ l c | c c }
            \toprule
            Supernet   & Pretrain (Pipeline) & $Cost$   & Top-1 (\%)    \\
            \midrule
            ResNet-50  & Y (\DMSp)           & $410+41$ & 73.8          \\
            \midrule
            ResNet-50  & N (\DMSnp)          & $0+41$   & 73.1          \\
            ResNet-101 & N (\DMSnp)          & $0+79$   & 74.2          \\
            ResNet-152 & N (\DMSnp)          & $0+116$  & \textbf{74.6} \\
            \bottomrule
        \end{tabular}
    \end{small}
    \vskip -0.1in

    \label{tab:initial}
\end{table}

It's widely accepted that both pretraining and increasing supernet size can improve the performance of searched models \cite{liu2018rethinking,frankle2018lottery}. However, both of them also introduce more resource consumption. Here, we compare the influence of conducting pretraining and increasing supernet size on performance and resource consumption.
Note, when we load pretrained weights, it means we use ``\DMSp'' pipeline, and when we do not load pretrained weights, it means we use ``\DMSnp'' pipeline.
We search for ResNet models of 1G MACs models with different supernet sizes and compare their performance and resource consumption. The results are shown in Table \ref{tab:initial}.

For pretraining, searching on a pretrained ResNet-50 surpluses searching on a randomly initialized ResNet-50 by 0.7\%. However, pretraining also increases about 10 times the resource consumption.
For supernet sizes, searching on a randomly initialized ResNet-152 surpluses searching on a randomly initialized ResNet-50 by 1.5\%, and only increases about 3 times the resource consumption.
We find increasing supernet size is more efficient than using strong pretrained weights for our method.
Therefore, our default pipeline ``\DMSnp'' just searches on randomly initialized supernets, making our method much more efficient than high-search-cost NAS methods.
It also makes our method easier to be applied in the real world.
This benefit originates from the high search efficiency of our method, making us use much fewer epochs to search a model than that of pretraining a supernet.

\subsubsection{Search with Latency Constraint}
\label{sec:abla_latency}

Except for the MAC constraint, we also conduct experiments with \emph{latency constraint}. We utilize the supernet of OFA \cite{cai2019once} and search for models under the latency constraint of 3ms on RTX3090. The results are shown in Table \ref{tab:latency}.
Our method costs much less search cost and achieves the same performance as the evolutionary algorithm, again.
It demonstrates that our method is applicable under different constraints.
The implementation detail of our latency constraint is detailed in \hyperref[sec:constrait]{Appendix A.1.2}.

\begin{table}[t]
    \centering
    \caption{Experiment with Latency Constraint: We compare our DMS with an evolutionary algorithm (EA) under the latency constraint of 3ms on RTX3090.
        Only EA + predictor needs a public cost, which is used to train an accuracy predictor.
        The unit for search cost is GPU hours, respectively.
        We use the pipeline of ``\DMSpn'' in this table.
    }

    \vskip 0.1in
    \begin{small}
        \begin{tabular}{ l | c c  c}
            \toprule
            Method        & Top-1 (\%) & Latency (ms) & Search Cost       \\
            \midrule
            \thead[l]{EA + predictor                                      \\ \cite{cai2019once}}  & 78.3       & 3            & 40 + 0                  \\
            \DMSpn (ours) & 78.3       & 3            & \textbf{0 + 0.05} \\
            \bottomrule
        \end{tabular}

    \end{small}
    \vskip -0.1in

    \label{tab:latency}
\end{table}

\subsubsection{Ablation Study of Search Time}
\label{sec:abla_search_time}
\begin{table}[t]
    \caption{Ablation Study on Search Time. We use the pipeline of ``\DMSnp'', which does not load pretrained weights, in this table.}

    \begin{center}
        \vskip 0.1in
        \begin{small}
            \begin{tabular}{ l | c c }
                \toprule
                Search Time & MACs (G) & Top-1 (\%)    \\
                \midrule
                ResNet-18   & 1.8      & 69.9          \\
                \midrule
                3 epochs    & 1        & 71.6          \\
                5 epochs    & 1        & 72.9          \\
                10 epochs   & 1        & \textbf{73.1} \\
                20 epochs   & 1        & 72.8          \\
                \bottomrule
            \end{tabular}
        \end{small}
        \vskip -0.1in
    \end{center}
    \label{tab:time}
\end{table}

We assessed the relationship between search time and final performance, as detailed in Table \ref{tab:time}. Our method achieves the best performance with 10 epochs. It's about 1/10 of the epochs of retraining the searched model. This proves the high search efficiency of our method, where a few epochs are enough to search for a model.
Besides, even though when the search time is only 3 epochs, the performance of the searched model is still better than the human-designed ResNet-18 with fewer MACs.

\subsubsection{Ablation Study of Hyperparameters for our Method}
\label{sec:abla_hyper}

\begin{table}[t]
    \caption{Ablation Study on Unfixed Hyperparameters. "/" denotes the model is not able to reach our resource target. We use the pipeline of ``\DMSnp'', which does not load pretrained weights, in this table.}

    \begin{center}
        \vskip 0.1in
        \begin{small}
            \begin{tabular}{ l | c c c c }
                \toprule
                \backslashbox{$\lambda_{resource}$}{$lr_{structure}$} & 5e-2 & 5e-3          & 5e-4 & 5e-5 \\
                \midrule
                0.1                                                   & /    & /             & /    & /    \\
                1                                                     & 73.0 & \textbf{73.1} & 72.9 & /    \\
                10                                                    & 72.5 & 72.2          & 72.6 & 70.9 \\
                \bottomrule
            \end{tabular}
        \end{small}
        \vskip -0.1in
    \end{center}
    \label{tab: hyper}
\end{table}

We divide the hyperparameters of our method into two categories: fixed hyperparameters and unfixed hyperparameters. Fixed hyperparameters are hyperparameters that are fixed for all models, while unfixed hyperparameters are hyperparameters that needs to be turned for different models.

The fixed hyperparameters include the decay rate for Taylor importance and the temperature $\lambda$ for our differentiable topk operator.

Taylor importance \cite{molchanov2019tayler} is a well-known method to measure the importance of elements, and the decay of moving average is also widely used in the literature. Therefore, we directly use the decay rate of 0.99 regarding prior works.

Temperature $\lambda$ of our differentiable topk. The temperature is used to polarize \cite{zhuang2020polar} the mask of elements. Directly selecting a value that can polarize the mask of elements is enough. Thanks to our importance normalization, the temperature can be directly computed by closed-form, detailed in Section \ref{sec:mask}. The temperature $\lambda$ is set to $N$ for width elements and $4N$ for depth elements and the number of heads in attention mechanisms. $N$ is the number of elements in the corresponding dimension. They work well for all models.

Therefore, we do not conduct an ablation study on these fixed hyperparameters.

The unfixed hyperparameters include the weight of resource constraint loss $\lambda_{resource}$ and the learning rate for structure parameters $lr_{structure}$. They are used to control the update of the structure parameters. The update value of a structure parameter is computed by $lr_{structure}\times (g_{task} + \lambda_{resource}\times g_{resource})$, where $g_{task}$ and $g_{resource}$ is the gradient of structure parameters with respect to the task loss and resource constraint loss,
Table \ref{tab: hyper} shows the ablation study results.

Obviously,
1) Smaller $\lambda_{resource}$ is better, as far as the model can reach the target resource constraint. Smaller $\lambda_{resource}$ means that the task loss takes more control of the update of the structure parameter.
2) When $\lambda_{resource}$ is small, the model is not sensitive to the change of $lr_{structure}$. When $\lambda_{resource}$ is large, a relatively large $lr_{structure}$ is better. This is because reaching the target resource constraint quickly can reduce the influence of the resource constraint loss, as resource constraint loss is zero when the model reaches the target resource constraint.

Therefore, the setting of $\lambda_{resource}$ and $lr_{structure}$ is not difficult. We first fix $lr_{structure}$ and turn $\lambda_{resource}$ to a small value and ensure the model can reach the target resource constraint. Then, we turn $lr_{structure}$ to a relatively large value, which makes the model reach the target resource constraint in the first hundreds of iterations. Only observing the resource decrease in the first epoch is enough to set these two hyperparameters.

Compared with other NAS methods, our method uses fewer hyperparameters. For example, ModelAmplification \cite{liu2022modelamplification} must turn at least five hyperparameters for different tasks and models.

\subsubsection{Ablation Study on Search Spaces}
\label{sec:abla_search_space}

In prior experiments, except for our own search space, we also apply our method on the search space defined by OFA \cite{cai2019once} in \hyperref[sec:evolution]{Section 4.1.2} and \hyperref[sec:abla_latency]{Appendix A.3.3}. Experiment results demonstrate our method works fine on these search spaces. Here, we further conduct an ablation study about search spaces.

There are two differences between the search space of our method and that of prior NAS methods.
\begin{itemize}
    \item On CNN models, we only search for width and depth, while prior methods usually also search for resolution and kernel size. We do not search for resolution and kernel size because we want to build a general search method that can be applied to various tasks and architectures. Searching for resolution and kernel size is not necessary for transformers and other tasks except for vision tasks.
    \item Due to our high search efficiency, our search space is more fine-grained, while prior methods usually implement a coarse-grained search space, containing much fewer sub-networks than ours. Besides, Coarse-grained search spaces also always need more expert knowledge to design, making it hard to build a general search method. Specifically, for a structure hyperparameter $x$, we directly search it in the range of $[1, x_{max}]$ with step 1, while most prior methods usually search it in the range of $[x_{min}, x_{max}]$ with a minimal step, such as 32 and 64.
\end{itemize}

Here, we conduct an ablation study based on Autoformer \cite{chen2021autoformer} search space. Autoformer implements a coarse-grained search space, such as searching embedding size with a minimal size of 192, a maximal size of 240, and a step of 24. We search two models with the exact same coarse-grained search space as Autoformer and a corresponding fine-grained search space. The results are shown in Table \ref{tab:auto}.

On the same supernet size, no matter the fine-grained search space or the coarse-grained search space, our method achieves better performance than Autoformer with less than 1/10 search costs. It proves our high search efficiency.
Besides, we find coarse-grained search yields better performance than fine-grained search space for our methods. This is because human-designed coarse-grained search space has been a pretty good sub-space of the fine-grained search space by the search conducted by human experts. The coarse-grained search space was updated by many human experts in different research projects, the ``search costs'' of this ``human searching stage'' may take a lot of time and resources.

Besides, we find increasing the supernet size makes the fine-grained search space achieve comparable performance with the coarse-grained search space. It demonstrates bigger supernet size contains better subnets, which can be searched by our method.

In this paper, as we want to build a general search method with the least human labor, we still use the fine-grained search space as our default search space.

\begin{table}[t]
    \caption{Ablation Study on Search Space. We compare the performance with Autoformer with the exact same search space. We use the pipeline of ``\DMSnp'', which does not load pretrained weights, in this table.}

    \begin{center}
        \vskip 0.1in
        \begin{small}
            \begin{tabular}{ l | c c c c }
                \toprule
                Method                                 & Search Space  & Supernet Size (G) & Top-1 (G)     & Search Cost (GPU days) \\
                \midrule
                Autoformer-T \cite{chen2021autoformer} & Carse-Grained & 2.2               & 74.7          & $>$ 25                 \\
                \midrule
                \DMSnp (ours)                          & Fine-Grained  & 2.2               & 74.8          & 2                      \\
                \DMSnp (ours)                          & Fine-Grained  & 6.1               & 75.1          & 5.5                    \\
                \DMSnp (ours)                          & Carse-Grained & 2.2               & \textbf{75.2} & 2                      \\
                \bottomrule
            \end{tabular}
        \end{small}
        \vskip -0.1in
    \end{center}
    \label{tab:auto}
\end{table}

\subsection{Implementation Details}

\subsubsection{Detail of Training Setting}
\label{sec:setting}

In general, given a baseline model and a training setting, we enlarge the baseline model as our supernet and decrease the number of epochs of the training setting as our searching setting. We list details of our experiment setting as shown below.

\textbf{EfficientNet}:
For all \DMSnp-ES variants, we pruned the supernets over a span of 30 epochs. For those \DMSnp-ES variants with MACs fewer than 0.5G, the pruning was conducted from EfficientNet-B4, using an input size of 224. Meanwhile, for \DMSnp-EN-B1 and B2, the pruning was initiated from EfficientNet-B7. The input sizes for \DMSnp-EN-B1 and B2 were 256 and 288, respectively.
Subsequently, the \DMSnp-EN variants were retrained using the corresponding training scripts of EfficientNet available in the Timm library \cite{rw2019timm}.

\textbf{ResNet}:
We pruned the ResNet over ten epochs, starting from the ResNet-152 model. After pruning, the ResNet was retrained utilizing the MMPretrain \cite{2023mmpretrain} training settings. This encompasses the foundational setting with a step learning scheduler and the rsb-a1 configuration.

\textbf{MobileNetV2}:
To search for the ideal structure for MobileNet, we commenced by enlarging MobileNetV2 before pruning. Specifically, all channel numbers were expanded by 1.5 times, and the number of blocks in each stage was doubled. The pruning process for MobileNetV2 spans 30 epochs. Subsequent to this, the architecture was retrained employing the MMPretrain training settings.

\textbf{Deit}:
We enhanced the depth of the Deit-small model, moving from 12 to 16, to serve as the supernet. The pruning for Deit was conducted with 30 epochs, including 20 epochs as a warmup phase. After pruning, we retrained the model using MMPretrain combined with the Swin training setting.

\textbf{Yolo-v8}
We used Yolo-v8 with deepen factor of 0.5 and widen factor of 0.5 as our supernet, while the original Yolo-v8-n has deepen factor of 0.33 and widen factor of 0.26. We used the training setting of Yolo-v8-n to train the supernet and pruned it over 30 epochs. The experiment of Yolo-v8 was conducted based on MMYolo \cite{mmyolo2022}.

\subsubsection{Detail of Search Cost Estimation}
\label{sec:ap_cost}

In this section, we delve into the specifics of how we estimate the search costs for other NAS methods as outlined in Table \ref{tab:eff}. The search cost of a searched model is divided into two parts: the public part and the private part. The public part is conducted for all sub-models, while the private part pertains to a specific sub-model.

\textbf{EfficientNet}: EfficientNet searches for common scaling strategies across all variants, thus incurring no private search cost. The public search cost estimate for EfficientNet is sourced directly from the ScaleNet paper \cite{xie2022scalenet}.

\textbf{ScaleNet} \cite{xie2022scalenet}: The ScaleNet paper explicitly presented their search cost, which includes a public cost of 379 GPU days and a private cost of 106 GPU days for several sub-models, totaling 21G MACs. We compute the private search cost for a sub-model based on the ratio of its MACs to the overall 21G MACs.

\textbf{ModelAmplification} \cite{liu2022modelamplification}: As a multi-shot NAS method, ModelAmplification requires training multiple models. For all sub-models, it utilizes a public proxy dataset and a proxy training script. Approximately 2007 epochs are expended to examine the proxy dataset, and an additional 2963 epochs are used for the proxy training script, leading to a total of 4970 epochs.
During the model search phase, for a variant with 390M MACs, ModelAmplification trains about 390 models per iteration. Assuming a ten-fold iteration search per model, this results in roughly 3000 epochs. By benchmarking the training time of EfficientNet-B0 on A100, we determine that 100 epochs require about 2.5 GPU days. As a result, the public search cost for ModelAmplification is at least 144 GPU days, while the private cost for the 390M MACs variant is 75 GPU days. We linearly scale the search costs of different variants based on their MACs.

\textbf{JointPruning} \cite{guo2021jointpruning}: As a gradient-based pruning method, JointPruning presumably employs a supernet and training script analogous to ours. We deduce its search cost based on the number of pruning epochs. JointPruning paper indicates that a quarter of the total training epochs is earmarked for model searching. In contrast, we utilize at most a tenth of the total epochs for this purpose. Hence, the search cost for JointPruning is 2.5 times that of ours.

\subsection{Visualization of Searched Model Structure}

\label{sec:visual}
\begin{figure}[t]
    \begin{center}
        \includegraphics[height=0.9\textheight]{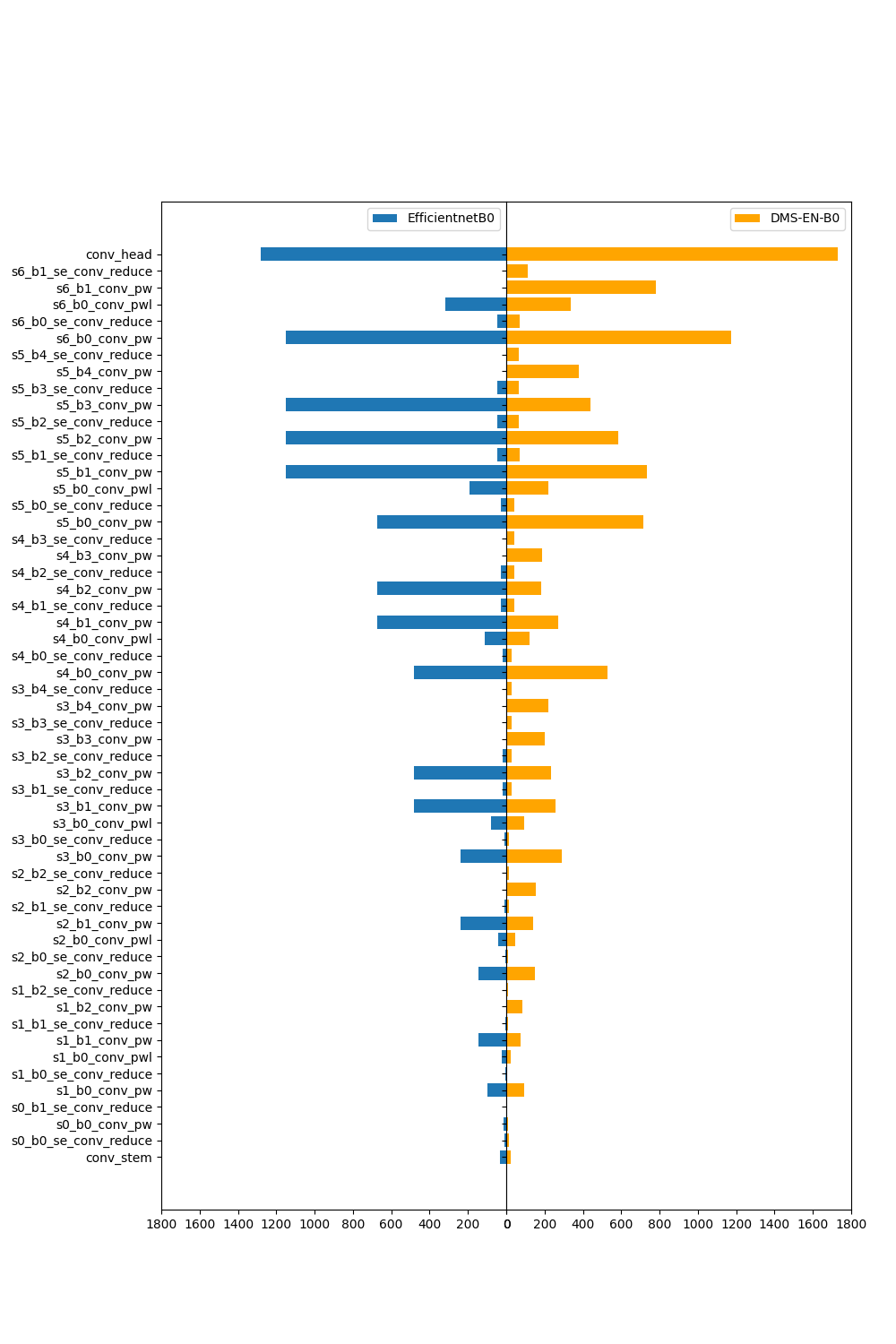}
        \caption{Visualization of Our Searched Structure. The x-axis represents the layers' width (channels/features), while the y-axis represents the layers. As \DMSnp-EN-B0 has more layers than EfficientNet-B0, the width of extra layers for EfficientNet-B0 are seen as 0.}
        \label{fig:structure}
    \end{center}
\end{figure}

In Figure \ref{fig:structure}, a visualization is provided to delineate the structural intricacies of our searched \DMSnp-EN-B0 in comparison to EfficientNet-B0.
A distinct observation that stands out is the depth of our \DMSnp-EN-B0. It possesses 8 more inverted residual blocks than its EfficientNet counterpart.
Furthermore, when we delve deeper into the channel distribution across different stages, it becomes evident that our \DMSnp-EN-B0 has undergone significant structural modifications, veering away from the traditional blueprint of EfficientNet-B0.
Such distinct differences underscore the fine-grained adaptability of our method, emphasizing its capability to recalibrate and refine models in a way that they are acutely tailored to the task.

\end{document}

%% file: math_commands.tex
\usepackage{amsmath,amsfonts,bm}

\def\eqref#1{equation~\ref{#1}}

\def\1{\bm{1}}

\def\vc{{\bm{c}}}

\def\vm{{\bm{m}}}

\DeclareMathAlphabet{\mathsfit}{\encodingdefault}{\sfdefault}{m}{sl}
\SetMathAlphabet{\mathsfit}{bold}{\encodingdefault}{\sfdefault}{bx}{n}